\documentclass[runningheads]{llncs}
\usepackage{graphicx}

\usepackage{tikz}
\usepackage{comment}
\usepackage{amsmath,amssymb} 
\usepackage{color}
\usepackage{graphicx}
\usepackage{booktabs}
\usepackage{multirow}
\usepackage[export]{adjustbox}
\usepackage{threeparttable}
\usepackage{subfloat}
\usepackage{tabularx}
\usepackage{adjustbox}
\usepackage{subcaption}
\usepackage[misc]{ifsym} 
\definecolor{linkcolor}{RGB}{255,0,0}
\definecolor{urlcolor}{RGB}{255,105,180}
\definecolor{citecolor}{RGB}{66,168,235}
\usepackage{hyperref}
\hypersetup{colorlinks=true,linkcolor=linkcolor,urlcolor=urlcolor,citecolor=citecolor}

\usepackage[accsupp]{axessibility}  

\begin{document}
\pagestyle{headings}
\mainmatter
\def\ECCVSubNumber{5532}  

\title{Panoptic-PartFormer: Learning a Unified model for Panoptic Part Segmentation} 

\titlerunning{Panoptic-PartFormer}
%
\author{Xiangtai Li\inst{1} \quad
Shilin Xu\inst{1} \quad
Yibo Yang\inst{1,3} \quad  \\
Guangliang Cheng\inst{2 \textrm{\Letter}} \quad
Yunhai Tong\inst{1 \textrm{\Letter}} \quad
Dacheng Tao\inst{3}}
\authorrunning{X. Li et al.}
\institute{\small Key Laboratory of Machine Perception, MOE, School of Artificial Intelligence, Peking University \\ \and SenseTime Research \and JD Explore Academy \\
\email{ lxtpku@pku.edu.cu, xushilin@stu.pku.edu.cn,chengguangliang@sensetime.com}}
\maketitle

\begin{abstract}
Panoptic Part Segmentation (PPS) aims to unify panoptic segmentation and part segmentation into one task. Previous work mainly utilizes separated approaches to handle thing, stuff, and part predictions individually without performing any shared computation and task association. In this work, we aim to unify these tasks at the architectural level, designing the first end-to-end unified method named Panoptic-PartFormer. In particular, motivated by the recent progress in Vision Transformer, we model things, stuff, and part as object queries and directly learn to optimize the all three predictions as unified mask prediction and classification problem. We design a decoupled decoder to generate part feature and thing/stuff feature respectively. Then we propose to utilize all the queries and corresponding features to perform reasoning jointly and iteratively. The final mask can be obtained via inner product between queries and the corresponding features. The extensive ablation studies and analysis prove the effectiveness of our framework. Our Panoptic-PartFormer achieves the new state-of-the-art results on both Cityscapes PPS and Pascal Context PPS datasets with around 70\% GFlops and 50\% parameters decrease. Given its effectiveness and conceptual simplicity, we hope the Panoptic-PartFormer can serve as a strong baseline and aid future research in PPS. Our code and models will be available at \url{https://github.com/lxtGH/Panoptic-PartFormer}.
\end{abstract}

\section{Introduction}
\label{sec:intro}

\begin{figure*}[h]
	\centering
	\includegraphics[width=0.90\linewidth]{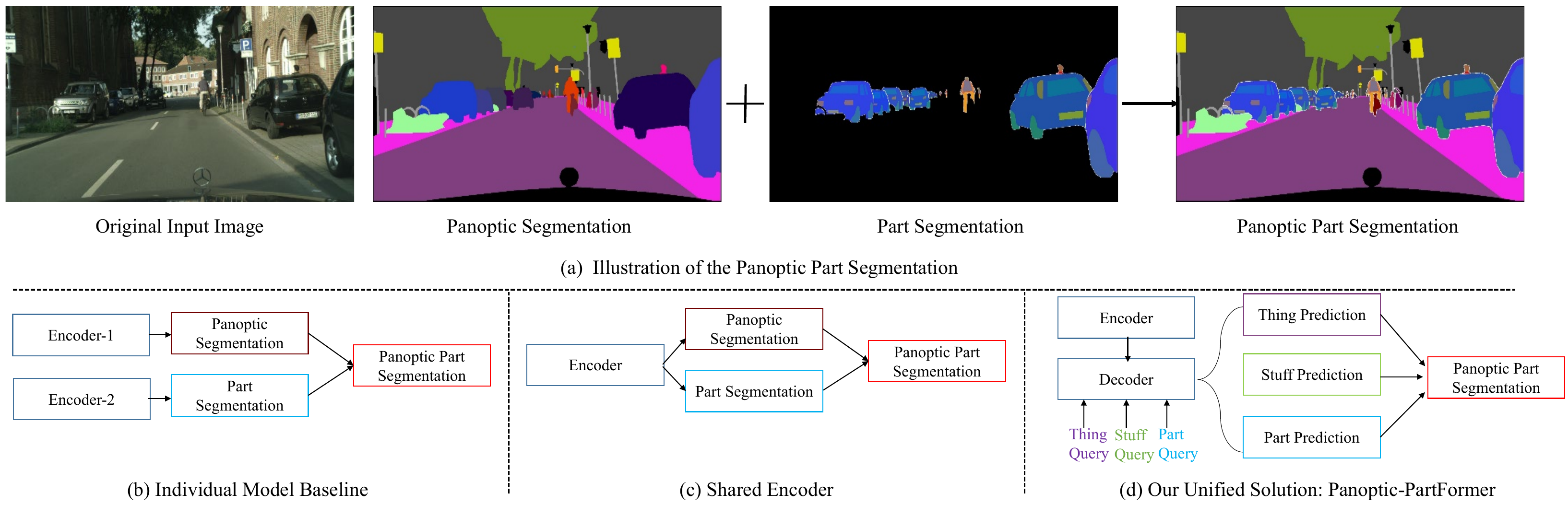}
	\caption{\small Illustration of Panoptic Part Segmentation (PPS) and different Approaches for solving such PPS task. (a) An illustration of the Panoptic Part Segmentation task. It combines the Panoptic Segmentation and Part Segmentation in a unified manner that provides the multi-level concept understanding of the scene
	(b) The baseline method proposed in~\cite{degeus2021panopticparts} combines results of panoptic segmentation and part segmentation. (c) Panoptic-FPN-like baseline~\cite{kirillov2019panopticfpn,li2019attention,xiong2019upsnet} adds part segmentation into the current panoptic segmentation frameworks. (d) Our proposed approach represents things, stuff, and part via object queries and performs joint learning. Best view in color.}
	\label{fig:teaser_01}
\end{figure*}

One essential problem in computer vision is to understand a scene at multiple levels of concept. In particular, when people perceive a scene, they can catch each visual entities such as car, bus, or person, and they can also understand the parts of entities, such as person-head and car-wheel, etc. The former is named as scene parsing, while the latter is termed as part parsing. One representative direction of unified scene parsing is Panoptic Segmentation (PS)~\cite{kirillov2019panoptic,kirillov2019panopticfpn}. It predicts a class label and an instance ID for each pixel. The part parsing has a wide range of definitions, such as human part or car part~\cite{liang2015human,geng2021part}. Both directions are independent, while both are equally important for many vision systems, including auto-driving and robot navigation~\cite{cordts2016cityscapes}.

Recently, the Panoptic Part Segmentation (PPS, or called Part-aware Panoptic Segmentation)~\cite{degeus2021panopticparts} is proposed to unify these multiple levels of abstraction into one single task. As shown in Fig.~\ref{fig:teaser_01}(a), it requires a model to output per-pixel scene-level classification for background stuff, segment things into individual instances and segment each instance into specific parts. Several baselines using combined approaches~\cite{maskrcnn,zhao2017pyramid,Zhao2019BSANet} were proposed to tackle this task. As shown in Fig.~\ref{fig:teaser_01}(a), they fuse different individual model predictions to obtain PPS results. In particular, they fuse the panoptic segmentation and part segmentation results from the non-shared backbone networks. This makes the entire process exceptionally complex with huge engineering efforts. Also, the shared computation and task association are ignored, which leads to inferior results. Another solution for this task is to make the part segmentation as an extra head with shared backbone as shown in Fig.~\ref{fig:teaser_01}(b). Such design is well explored in PS studies~\cite{xiong2019upsnet,kirillov2019panopticfpn,li2020panopticFCN,chen2020banet,li2018learning,porzi2019seamless,yang2019sognet,Wu2020AutoPanopticCM,cheng2020panoptic,axialDeeplab}. However, most of them treat PS as separated tasks~\cite{xiong2019upsnet} or sequential tasks with several post processing components~\cite{cheng2020panoptic}.

Since Detection Transformer (DETR)~\cite{detr}, there are several works~\cite{wang2020maxDeeplab,cheng2021maskformer,zhang2021knet} unifying both thing and stuff learning via \textit{object queries} in PS, which makes the entire pipeline elegant and achieves strong results where the mask classification and prediction can be performed directly. These results show that many complex components including NMS and box detection can be removed. In particular, such design considers the full scene understanding via performing interactions among things, stuff, and part simultaneously. Joint training with things, stuff, and part leads to better part segmentation results since the full scene information renders part representation a more discriminative information such as global context. Motivated by these analysis, we take a further step on the more challenging PPS task and propose the \textbf{first unified model} for this task.

In this paper, we present a simple yet effective baseline named Panopic-PartFormer, a unified model for PPS task. As shown in Fig.~\ref{fig:teaser_01}(c), we introduce three different types of queries for modeling thing, stuff, and part prediction, respectively. Then a decoupled decoder is proposed to generate fine-grained features. These features are used to decode thing, stuff, and part mask prediction. The decoupled decoder contains a part decoder and a scene decoder. For part decoder, we design a feature aligned decoder to keep more fine details in part. 
Rather than directly using the pixel-level self-attention in Transformer, we consider the recent works~\cite{peize2020sparse,zhang2021knet} that perform self-attention on query level. To be more specific, we focus on refining queries via their corresponding query features. We define the \textit{query feature} as \textit{grouped features} that are generated from the \textit{corresponding mask} of each query and \textit{decoder features}. The masks are generated via dot product between queries and decoder features. In particular, the initial query features are grouped via the initial mask prediction from the decoupled decoder. Then we perform updating object queries with the query features. This operation is implemented with one dynamic convolution~\cite{zhang2021knet,peize2020sparse} and multi-head self-attention layers~\cite{vaswani2017attention} between query and query features iteratively. The  former poses instance-wise information from feature to enhance the query learning where the parameters are generated by the features itself while the latter performs inner reasoning among different types of queries to model the relationship among thing, stuff, and part. Moreover, the entire procedure avoids pixel-level computation in other vision Transformer decoder~\cite{detr,cheng2021maskformer}. In this way, since all thing, stuff, and part information is encoded into query, the relation between these queries can be well explored and optimized jointly via pure mask based supervison. Extensive experiments (Sec.\ref{sec:experiment}) show that our approach achieves much better results than the previous design in Fig.~\ref{fig:teaser_01}(b).

Moreover, our Panopic-PartFormer can support both CNN backbones~\cite{resnet} and Transformer backbones~\cite{liu2021swin} for feature extraction. Panopic-PartFormer is also memory and computation efficient, which is mainly benefited by avoiding pixel-level computation of self-attention layers. Panopic-PartFormer can directly output thing, stuff, and part segmentation predictions in box-free and NMS-free manner. It can also be evaluated by sub-task of PPS such as Panoptic Segmentation. In the experiment part, we verify our panoptic segmentation predictions on Cityscaeps datasets~\cite{cordts2016cityscapes} and it also achieves better results than the recent works~\cite{li2020panopticFCN}. To sum up, our main contributions are as follows:
\begin{itemize}
	\item We present a novel, simple and effective baseline named Panopic-PartFormer for the PPS task. To the best of our knowledge, it is the first unified end-to-end model for this task. 
	
	\item We propose a decoupled decoder and a joint query updating and reasoning framework for the joint feature learning of thing, stuff, and part. Besides, a joint loss function is proposed to supervise the whole model.
	
	\item Extensive experiments and analyses indicate the effectiveness and generalization of our model. In particular, with our framework, we find the part segmentation can be improved significantly via joint training.  We achieve the new state-of-the-art results on two challenging PPS benchmarks including Pascal Context PPS dataset (about 6-7\% PartPQ gain on ResNet101, 10\% PartPQ gain using swin Transformer~\cite{liu2021swin}) and Cityscapes PPS dataset (about 1-2\% PartPQ gain). 
\end{itemize}
\section{Related Work}
\label{sec:relatedwork}


\noindent
\textbf{Part Segmentation.}
Most previous approaches for both instance and semantic part segmentation mainly focus on human analysis~\cite{qi2018learning,shen2019human}. Several works~\cite{fang2018weakly,liu2018cross,wang2019CNIF} design specific methods for semantic part segmentation which are in category-level settings. There are two paradigms for human part segmentation: \textit{top-down} pipelines~\cite{li2017holistic,yang2019parsing,ruan2019devil,ji2019learning,yang2020eccv} and \textit{bottom-up} pipelines~\cite{gong2018instance,li2017multiple,zhao2018understanding,zhou2021differentiable}. Meanwhile, there are also several works focusing on task-specific part segmentation~\cite{Zhao2019BSANet,lin2019face,michieli2020gmnet}. Compared with these methods, the focus of this paper is to solve the PPS task which naturally contains the part segmentation as a sub-task.

\noindent
\textbf{Panoptic Segmentation.} 
Earlier work~\cite{kirillov2019panoptic} mainly performs segmentation for things and stuff via separated networks where the original benchmark directly combines predictions of things and stuff from different models. To alleviate the computation cost, recent works~\cite{kirillov2019panopticfpn,li2019attention,chen2020banet,li2018learning,porzi2019seamless,yang2019sognet,Wu2020AutoPanopticCM} are proposed to model both stuff segmentation and thing segmentation in one model with different task heads. Detection based methods~\cite{xiong2019upsnet,kirillov2019panopticfpn,qiao2021detectors,hou2020real} usually represent things with the box prediction while several bottom-up models~\cite{cheng2020panoptic,yang2019deeperlab,gao2019ssap,axialDeeplab} perform grouping instance via pixel-level affinity or center heat maps from semantic segmentation results. The former introduces complex process while the latter suffers from the performance drops in complex scenarios.
Recently, several works~\cite{wang2020maxDeeplab,cheng2021maskformer,zhang2021knet} propose to  directly obtain segmentation masks without box supervision. However, these works do \textit{not} cover the \textit{knowledge of part-level semantics} of images which can provide more comprehensive information for scene understanding.

\noindent
\textbf{Panoptic Part Segmentation.}
To better understand the full scene, the PPS task~\cite{degeus2021panopticparts} is proposed. This work annotates two datasets (Cityscape PPS~\cite{cordts2016cityscapes} and Pascal Context PPS~\cite{Everingham2010Pascal}) and proposes a new metric named PartPQ~\cite{degeus2021panopticparts} for evaluation. This work also presents several baseline methods to obtain the final results. However, these methods are all separated networks for instance and semantic segmentation to obtain the panoptic segmentation or use existing panoptic segmentation algorithms with part semantic segmentation as an isolated sub-network (shown in Fig~.\ref{fig:teaser_01}(a) and (b)). For the comparison, our goal is to design a unified and effective network for all the tasks.

\noindent
\textbf{Vision Transformer.} There are mainly two different usages for Transformer in vision: feature extractor and query modeling. Compared with CNN, vision Transformers~\cite{VIT,liu2021swin,deit_vit} have more advantages in modeling global-range relation among the image patch features. The second design is to use the object query representation. DETR~\cite{detr} models the object detection task as a set prediction problem with learnable queries. The following works~\cite{peize2020sparse,zhu2020deformabledetr} explore the locality of the learning process to improve the performance of DETR. Query based learning can also be applied to other fields ~\cite{QueryInst,meinhardt2021trackformer}. Our methods are inspired by these works with the goal of unifying and simplifying the PPS task based on query learning.
\section{Method}
\label{sec:method}


\subsection{Basic Architecture}
\begin{figure*}[t]
	\centering
	\includegraphics[width=1.0\linewidth]{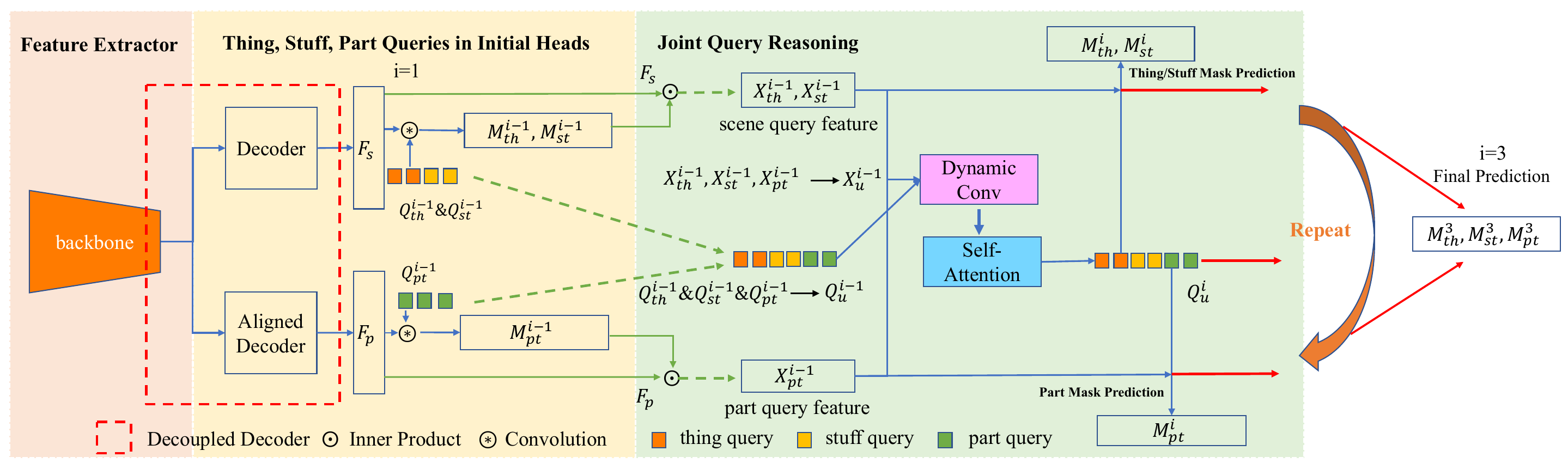}
	\caption{\small Our proposed Panoptic-PartFormer. It contains three parts: (1) a backbone to extract features (Red area) (b) a decoupled decoder to generate scene features and part features along with the initial prediction heads to generate initial mask predictions. (Yellow area) (c) a cascaded Transformer decoder to jointly do reasoning between the query and query features. (Green area) \textcolor{green}{Green arrows} mean input (come from previous stage) while \textcolor{red}{Red arrows} represent current stage output (used for next stage). $i=3$ is the final mask outputs.}
	\label{fig:methods}
\end{figure*}

Fig.~\ref{fig:methods} presents an overall illustration of the proposed method. 
Our method contains three parts: (1) an encoder backbone to extract features; (2) a decoupled decoder to obtain the scene features and part features individually. We denote the scene features are used to generate things, stuff masks while the part features are used to generate part masks; (3) a Transformer decoder which takes three different types of queries and backbone features as inputs and provides thing, stuff, and part mask predictions. 


\noindent
\textbf{Encoder network:}
We first extract image features for each input image. It contains a backbone network (Convolution Network~\cite{resnet} or Vision Transformer~\cite{liu2021swin}) with Feature Pyramid Network~\cite{fpn} as neck. This results in a set of multiscale features which are the inputs of the decoupled decoder. 

\noindent
\textbf{Decoupled decoder:} The decoupled decoder has two separate decoder networks to obtain features for scene feature and part feature, respectively. The former is used to decode both thing and stuff predictions, while the latter is applied to decode the part prediction. Our motivation is that part segmentation has different properties from panoptic segmentation. First, part features need a more precise location and fine details. Second, scene features focus on mask proposal level prediction while part features pay more attention to the inner parts of mask proposal, which conflicts with each other. We show that the decoupled design leads to better results in the experimental section. (see Sec.~\ref{sec:ablation}).

For implementation, we adopt semantic FPN design~\cite{kirillov2019panopticfpn} to fuse features in a top-down manner. Thus, we obtain the two features named $F_{s}$ and $F_{p}$. In particular, for part segmentation, we design a light-weight aligned feature decoder for part segmentation. Rather than the naive bilinear upsampling, we propose to learn feature flow~\cite{sfnet,zhou2021differentiable} to warp the low resolution feature to high resolution feature. Then we sum all the predictions into the highest resolution as semantic FPN. Moreover, to preserve more locational information, we add the positional embedding to each stage of the semantic FPN following the previous works~\cite{wang2020solo,wang2020solov2}. In summary, decoupled decoder outputs two separate features: scene features $F_{s}$ and part features $F_{p}$. The former is used to generate thing and stuff masks while the latter is used to generate part masks and both have the same resolution.

\noindent
\subsection{Thing, Stuff, and Part as Queries with Initial Head Prediction} 

Previous works~\cite{cheng2021maskformer,zhang2021knet,wang2020maxDeeplab} show that single mask classification and prediction can achieve the state-of-the-art results on COCO~\cite{coco_dataset}. Motivated by this, our model treats thing, stuff, and part as the input queries to directly obtain the final panoptic part segmentation. Following previous works~\cite{zhang2021knet,peize2020sparse}, the initial weights of these queries are directly obtained from the first stage weights of the initial decoupled decoder prediction. For mask predictions of thing, stuff, and part, we use three $1 \times 1$ convolution layers to obtain the initial outputs of thing, stuff, and part masks. These layers are appended at the end of the decoupled decoder.

All these predictions are directly supervised with corresponding ground truth masks. As shown in ~\cite{peize2020sparse,zhang2021knet}, using such initial heads can avoid heavier Transformer encoder layers for pixel-level computation, since the corresponding query features can be obtained via mask grouping from the initial mask prediction. 

In this way, we obtain the three different queries for thing, stuff, and part along with their initial mask prediction. We term them as $Q_{th}$, $Q_{st}$, $Q_{pt}$ and $M_{th}$, $M_{st}$, $M_{pt}$ with shapes $N_{th} \times d$, $N_{st} \times d$, $N_{pt} \times d$ and shapes $N_{th} \times H \times W$, $N_{st} \times H \times W$, $N_{pt} \times H \times W$. $d$, $W$, $H$ are the channel number, width, height of feature $F_{p}$ and $F_{s}$.  $N_{th}$, $N_{st}$, $N_{pt}$ are numbers of categories for thing, stuff, and part classification.

\noindent
\subsection{Joint Thing, Stuff, and Part Query Reasoning} 

The cascaded Transformer decoder takes previous mask predictions, previous object queries and decoupled features as inputs and outputs the current refined mask predictions and object queries. The refined mask predictions and object queries along with decoupled features will be the inputs of the next stage. The relationship between queries and query features is jointly learned and reasoned. 

Our key insights are: Firstly, joint learning can learn the full correlation between scene features and part features. For example, car parts must be on the road rather than in the sky. Secondly, joint reasoning can avoid several scene noisy cases, such as car parts on the human body or human parts on the car. We find joint learning leads to better results (see Sec.~\ref{sec:ablation}).

We combine the three queries and the three mask predictions into a unified query $Q_{u}^{i-1}$ and $M_{u}^{i-1}$ where $Q_{u}^{i-1} = concat(Q_{th}, Q_{st}, Q_{pt})$, $ Q_{u}^{i-1} \in \boldsymbol{R}^{(N_{th} + N_{st} + N_{pt}) \times d}$ and $M_{u}^{i-1} = concat(M_{th}^{i-1}, M_{st}^{i-1}, M_{pt}^{i-1})$, $ M_{u}^{i-1} \in \boldsymbol{R}^{(N_{th} + N_{st} + N_{pt}) \times H \times W}$. 
$i$ is the stage index of our Transformer decoder. $i=1$ means the predictions come from the initial heads. Otherwise, it means predictions from the outputs of previous stage.
$concat$ is preformed along the first dimension. 

In particular, following the previous work~\cite{zhang2021knet}, we first obtained query features $X^i$ via grouping from previous mask predictions $M_{u}^{i-1}$ and input features ($F_{p}$, $F_{s}$) shown in Equ.~\ref{equ:grouping} (dot product in Fig.~\ref{fig:methods}). We present this process in one formulation for simplicity.

\begin{equation}
    X^i = \sum_u^W\sum_v^H M^{i-1}(u, v) \cdot F(u, v),
 \label{equ:grouping}
\end{equation}
where $X^i \in \boldsymbol{R}^{(N_{th} + N_{st} + N_{pt}) \times d}$ is the per-instance extracted feature with the same shape as  $Q_{u}$, $M^{i-1}$ is the per-instance mask extracted from the previous stage $i-1$, and $F$ is the input feature extracted for the decoupled decoder head. $u$, $v$ are the indices of spatial location. $i$ is layer number and starts from 1. As shown in center part of Fig~\ref{fig:methods}, the part mask prediction and scene mask prediction are applied on corresponding features ($F_{p}$, $F_{s}$) individually where we obtain part query features $X_{pt}^i$ and scene query features $X_{th}^i$ and $X_{st}^i$. Then we combine these query features through $X_{u}^i = concat(X_{th}^i, X_{st}^i, X_{pt}^i)$. These inputs are shown in the green arrows in Fig.~\ref{fig:methods}.

Then we perform a dynamic convolution~\cite{tian2020conditional,zhang2021knet,peize2020sparse} to refine input queries $Q_{u}^{i-1}$ with the query features $X_{u}^i$ which are grouped from their masks.

\begin{equation}
    \hat{Q}_{u}^{i-1} = DynamicConv(X_{u}^{i}, Q_{u}^{i-1}),
 \label{equ:dynamic}
\end{equation}
where the dynamic convolution uses the query features $X_{u}^i$ to generate parameters to weight input queries $Q_{u}^{i-1}$. To be more specific, $DynamicConv$ uses input query features $X_{u}^{i}$ to generate gating parameters via MLP and multiply back to the original query input $Q_{u}^{i-1}$. 
Our motivation has two folds: Compared with pixel-wise MHSA~\cite{cheng2021maskformer,detr}, dynamic convolution introduces less computation and faster convergence for limited computation.
Secondly, it poses the instance-wised information to each query dynamically during training and inference, which shows better generalization and has complementary effects with MHSA. More details can be found in Sec.~\ref{sec:ablation}.

This operation absorbs more fine-grained information to help query look for more precise location. 

In particular, we adopt the same design~\cite{zhang2021knet} by learning gating functions to update the refined queries. The $DynamicConv$ operation is shown as follows:
\begin{equation}
    \hat{Q}_{u}^{i-1} = Gate_{x}(X_{u}^{i})X_{u}^{i} + Gate_{q}(X_{u}^{i}) Q_{u}^{i-1},
\end{equation}
where $Gate$ is implemented with a fully connected (FC) layers followed by LayerNorm (LN), and a sigmoid layer. We adopt two different gate functions including $Gate_{x}$ and $Gate_{q}$. The former is to weight the query features, while the latter is to weight corresponding queries.

After that, we adopt one self-attention layer with feed forward layers~\cite{vaswani2017attention,wang2020maxDeeplab} to learn the correspondence among each query and update them accordingly. This operation leads to the full correlation among queries, shown as follows:

\begin{equation}
    Q_{u}^{i} = FFN(MHSA(\hat{Q}_{u}^{i-1}) + \hat{Q}_{u}^{i-1}),
 \label{equ:selfattention}
\end{equation}
where $MHSA$ means Multi Head Self Attention, $FFN$ is the Feed Forward Network that is commonly used in current vision Transformers~\cite{detr,VIT}. The output refined query has the same shape as the input, i.e. $Q_{u}^{i} \in \boldsymbol{R}^{(N_{th} + N_{st} + N_{pt}) \times d}$.

Finally, the refined masks are obtained via dot product between the refined queries $Q_{u}^i$ and the input features $F_{p}$, $F_{s}$. For mask classification, we adopt several feed forward layers on $Q_{u}^{i}$ and directly output the class scores (thing, stuff, and part). For mask prediction, we also adopt several feed forward layers on $Q_{u}^{i}$ and then we perform the inner product between learned queries and features ($F_{s}$ $F_{p}$) to generate scene masks (thing and stuff) and part masks of stage $i$.
These masks will be used for the next stage input as shown in the red arrows in Fig~\ref{fig:methods}. The process of Equ.~\ref{equ:grouping}, Equ.~\ref{equ:dynamic} and Equ.~\ref{equ:selfattention} will be repeated for several times. We set the iteration number to 3 by default. The inter mask predictions are also optimized by mask supervision.

\noindent
\textbf{Discussion.} We admit that we use the dynamic convolution and self-attention among queries that are proposed by~\cite{peize2020sparse,zhang2021knet}. However, we \textbf{do not claim} this is our contribution
for Panoptic-PartFormer. Our main contribution is a system
level unified model for this challenging task(PPS) and we
are the first work to prove that joint learning of the thing,
stuff and part learning benefits PPS tasks than many other
designs. More details can be found in supplementary.

\noindent
\subsection{Training and inference} 

\noindent
\textbf{Training:} To train the Panoptic-PartFormer, we need to assign ground truth according to the pre-defined cost since all the outputs are encoded via queries. In particular, we mainly follow the design of~\cite{cheng2021maskformer} to use bipartite matching as a cost by considering both mask and classification results. After the bipartite matching, we apply a loss jointly considering mask prediction and classification for each thing, stuff, and part. In particular, we apply focal loss~\cite{focal_loss} on both classification and mask prediction. We also adopt dice loss~\cite{dice_loss} on mask predictions ($L_{part}$, $L_{thing}$, $L_{stuff}$). Such settings are \textit{the same as previous works}~\cite{detr,cheng2021maskformer}. The loss for each stage $i$ can be formulated as follows: 
\begin{equation}
    \mathcal{L}_{i} = \lambda_{part} \cdot \mathcal{L}_{part} + \lambda_{thing} \cdot \mathcal{L}_{thing} + \lambda_{stuff} \cdot \mathcal{L}_{stuff} +\lambda_{cls} \cdot \mathcal{L}_{cls}
\end{equation}
Note that the losses are applied to each stage $\mathcal{L}_{final} = \sum_{i}^N\mathcal{L}_i,$ where $N$ is the total stages applied to the framework. We adopt $N = 3$ and all $\lambda$s are set to 1 by default.

\noindent
\textbf{Inference:} We directly get the output masks from the corresponding queries according to their sorted scores. To obtain the final panoptic part segmentation, we first obtain the panoptic segmetnation results and then merge part masks into panoptic segmentation results. For panoptic segmentation results, we adopt the method used in Panoptic-FPN~\cite{kirillov2019panopticfpn} to merge panoptic mask. For part merging process, we follow the original PPS task to obtain the final panoptic part segmentation results. For scene-level semantic classes that do not have part classes, we simply copy the predictions from panoptic segmentation. For predicted instances with the part, we extract the part predictions for the pixels corresponding to this segment. Otherwise, if a part prediction contains a part class that does not correspond to the scene-level class, we set it to \textit{void} label. This setting mainly follows the previous work~\cite{degeus2021panopticparts}.
\section{Experiment}
\label{sec:experiment}


\begin{table*}[t]
\centering
\begin{adjustbox}{width=0.95\textwidth}
\begin{tabular}{ll||cc|ccc|c||ccc}
\bottomrule
 &  & \multicolumn{3}{c|}{\textbf{PQ}} & \multicolumn{3}{c}{\textbf{PartPQ}} \\ 
\textbf{Panoptic seg. method}  &  \textbf{Part seg. method}  & All & P & NP & All & P & NP \\ 
\midrule

\textit{\textbf{Cityscapes Panoptic Parts} validation set} \\
UPSNet \cite{xiong2019upsnet}(ResNet50) & DeepLabv3+ \cite{deeplabv3plus}(ResNet50)  & 59.1 & 57.3 & 59.7  & 55.1 & 42.3 & 59.7 \\
DeepLabv3+(ResNet50) \& Mask R-CNN(ResNet50) \cite{maskrcnn} & DeepLabv3+ \cite{deeplabv3plus} (Xception-
65)  & 61.0 & 58.7 & 61.9 &  56.9 & 43.0 & 61.9  \\
\hline
\textbf{Panoptic-PartFormer (ResNet50)}  & & \textbf{61.6} & \textbf{60.0} & \textbf{62.2} & \textbf{57.4} & \textbf{43.9} & \textbf{62.2} \\ 
\hline

EfficientPS \cite{mohan2021efficientps}(EfficientNet)~\cite{tan2019efficientnet} &  BSANet \cite{Zhao2019BSANet}(ResNet101)   & 65.0 & 64.2 & 65.2   & 60.2 & \textbf{46.1} & 65.2  \\
HRNet-OCR (HRNetv2-W48)~\cite{ocrnet,wang2020deep} \& PolyTransform~\cite{liang2020polytransform} &  BSANet \cite{Zhao2019BSANet}(ResNet101)  & 66.2 & 64.2 & 67.0  & 61.4 & 45.8 & 67.0 \\
\hline

\textbf{Panoptic-PartFormer (Swin-base)} & & \textbf{66.6}  & \textbf{65.1}  & \textbf{67.2}  & \textbf{61.9}  & \textbf{45.6} & \textbf{68.0}  \\

\bottomrule
\end{tabular}
\end{adjustbox}
\caption{\small \textbf{Experiment Results on CPP.} Previous works combine results from commonly used (top), and state-of-the-art methods (bottom) for semantic segmentation, instance segmentation, panoptic segmentation and part segmentation. Metrics split into \textit{P} and \textit{NP} are evaluated on scene-level classes with and without parts, respectively.}
\label{tab:experiments_res_city}
\end{table*}

\noindent
\textbf{Datasets.}
We mainly carry out experiments on two datasets including Cityscapes Panoptic Parts (CPP) and PASCAL Panoptic Parts (PPP), which are based on the established scene understanding datasets Cityscapes~\cite{cordts2016cityscapes} and PASCAL VOC~\cite{Everingham2010Pascal}, respectively. The CPP extends with part-level semantics the Cityscapes dataset~\cite{cordts2016cityscapes} and is annotated with 23 part-level semantic classes. In particular, 5 scene-level semantic classes from the {human} and {vehicle} high-level categories are annotated with parts. The CPP contains 2975 training and 500 validation images. PPP extends the PASCAL VOC 2010 benchmark~\cite{Everingham2010Pascal} with part-level and scene-level semantics. PPP has 4998 training and 5105 validation images. To perform fair comparison, following previous settings~\cite{degeus2021panopticparts,Everingham2010Pascal}, we perform experiments 59 scene-level classes (20 things, 39 stuff), and 57 part classes.  We further report Cityscapes Panoptic Segmentation validation set~\cite{cordts2016cityscapes} results as sub-task comparison.

\noindent
\textbf{Experiment Settings.} ResNet~\cite{resnet} and Swin Transformer~\cite{liu2021swin} are adopted as the backbone networks and other layers use Xavier initialization~\cite{xavier_init}. The optimizer is AdamW~\cite{ADAMW} with weight decay 0.0001. The training batch size is set to 16 and all models are trained with 8 GPUs. For PPP datasets, we first pretrain our model on COCO dataset~\cite{coco_dataset} since most previous baselines~\cite{degeus2021panopticparts} are pretrained on COCO. For PPP dataset, we adopt the multiscale training~\cite{detr} by resizing the input images from the scale 0.5 to scale 2.0. We also apply random crop augmentations during training, where the train images are cropped with probability 0.5. For CPP dataset, we follow the similar setting in Panoptic-Deeplab~\cite{cheng2020panoptic} where we resize the images with scale rang from 0.5 to 2.0 and randomly crop the whole image during training with batch size 16. All the results are obtained via single scale inference. 

\noindent
\textbf{Metric.}
We report PartPQ~\cite{degeus2021panopticparts} and PQ~\cite{kirillov2019panoptic} as the main metrics, since PartPQ is a unified metric that contain both scene-level output and part-level output. Part segmentation results such as mIoU can be found in the appendix file. The PartPQ per scene-level class $l$ is formalized as $\textrm{PartPQ} = \frac{\sum_{(p,g) \in \textit{TP}}\textrm{IOU\textsubscript{p}}(p,g)}{|\textit{TP}| + \frac{1}{2}|\textit{FP}|+ \frac{1}{2}|\textit{FN}|}$. ${TP}$ is true positive, ${FP}$ is false positive, and ${FN}$ is false negative segments, receptively. The definition of these is based on the Intersection Over Union (IOU) between a predicted segment $p$ and a ground-truth segment $g$ for a class $l$ (where $l \in \mathcal{L}$, $\mathcal{L}$ is the label set). A prediction is a ${TP}$ if it has an overlap with a ground-truth segment with an $\textrm{IOU} > 0.5$. An ${FP}$ is a predicted segment that is not matched with the ground-truth, and an ${FN}$ is a ground-truth segment not matched with a prediction. $IOU$ contains two cases (part and non-part): 
$$ \textrm{IOU\textsubscript{p}}(p,g) =
    \begin{cases}
      \textrm{mean IOU\textsubscript{part}}(p,g), & \textrm{$l \in \mathcal{L}^\text{parts}$}\\
      \textrm{IOU\textsubscript{inst}}(p,g), & \textrm{$l \in \mathcal{L}^\text{no-parts}$}
    \end{cases} $$

\subsection{Main Results}

\noindent
\textbf{Results on Cityscape Panoptic Part Dataset.} In Tab.~\ref{tab:experiments_res_city}, we compare our Panoptic-PartFormer with previous baselines. All the models use the single scale inference without test time augmentation. Our method with ResNe50 backbone achieves 57.4\% PartPQ which outperforms the previous work using complex pipelines~\cite{deeplabv3plus,maskrcnn} with even stronger backbone~\cite{chollet2017xception}. For the same backbone, our method results in 2.3\% PartPQ gain over the previous baseline. 
For large model comparison, our method with Swin-Transformer achieves 61.9 \%PartPQ. It outperforms previous works that use state-of-the-art individual models~\cite{ocrnet,wang2020deep,Zhao2019BSANet,liang2020polytransform} by 0.5\%. Note that the best model from HRNet~\cite{ocrnet} is pretrained using Mapillary dataset~\cite{neuhold2017mapillary}.
We follow the same pipeline for fair comparison. Both settings prove the effectiveness of our approaches.

\noindent
\textbf{GFlops and Parameter Comparison.} Since our method is one single unified model, our Panoptic-PartFormer has advantages on both GFlops and Parameters. Since the work~\cite{liang2020polytransform} is not public available, we estimate the lower bound by its baseline model~\cite{maskrcnn}.
As shown in Tab.~\ref{tab:cpp_detailed}, our model obtain around 60\% GFlops drop and 70\% parameter drop.

\noindent
\textbf{Results on Pascal Panoptic Part Dataset.} We further compare our method with previous work on the Pascal Panoptic Part dataset in Tab.~\ref{tab:experiments_res_ppp}. For different settings, our methods achieve state-of-the-art results on both PQ and PartPQ with a very significant gain. For ResNet backbone, our methods achieve \textbf{6-7\% gains} on PartPQ. Moreover, our resnet101 model can achieve better results than previous work using \textit{larger} backbone~\cite{qiao2021detectors}.
Using Swin Transformer base as backbone~\cite{liu2021swin}, our method achieves \textbf{47.4\% PartPQ} which shows the generalization ability on large model.

\noindent
\textbf{Results on Cityscapes Panoptic Segmentation.}
We also compare our method with several previous works on cityscapes panoptic validation set. As shown in Tab.~\ref{tab:experiments_res_cityscapes}, our Panoptic-PartFormer also achieves state-of-the-art results compared with previous works~\cite{li2020panopticFCN,cheng2020panoptic}. This proves the generalization ability of our framework. 

\begin{table}[!t]\setlength{\tabcolsep}{6pt}
	\centering
	\begin{threeparttable}
		\scalebox{0.60}{
			\begin{tabular}{l c c c c}
				\toprule[0.15em]
			Method &  PQ & PartPQ  & Param(M) & GFlops \\
				\toprule[0.15em]
		UPSNet + DeepLabv3+ (ResNet50) & 59.1 & 55.1 & $>$87 &  $>$890 \\ 
		\textbf{Panoptic-PartFormer (ResNet50)} & \textbf{61.6} & \textbf{57.4}  & \textbf{37.35} & \textbf{185.84}  \\ 
		\hline
		HRNet(OCR) +PolyTransform + BSANet & 66.2 & 61.4 & $>$181 & $>$1154 \\
        \textbf{Panoptic-PartFormer (Swin-base)} & \textbf{66.6}  & \textbf{61.9}  & \textbf{100.32} & \textbf{408.52} \\
	\bottomrule[0.1em]
	\end{tabular}}
		\caption{\small More detailed comparison on Cityscapes PPS dataset. GFlops are measured with 1200 $\times$ 800 input.}
		\label{tab:cpp_detailed}
	\end{threeparttable}
\end{table}

\begin{table}[t]
\centering
\begin{adjustbox}{width=0.50\textwidth}
\begin{tabular}{ll||cc}
\toprule[0.15em]
\textbf{Panoptic seg. method}  &  \textbf{Part seg. method}  & PQ & PartPQ \\ 
\midrule
\textit{\textbf{Pascal Panoptic Parts} validation set} \\
DeepLabv3+ \& Mask R-CNN \cite{maskrcnn}(ResNet50) & DeepLabv3+ \cite{deeplabv3plus}(ResNet50) & 35.0  &  31.4   \\ 
DLv3-ResNeSt269~\cite{zhang2020resnest} \& DetectoRS \cite{qiao2021detectors} &  BSANet \cite{Zhao2019BSANet}  & 42.0  & 38.3  \\ \hline
\textbf{Our Unified Approach} \\
\textbf{Panoptic PartFormer (ResNet50)} &  & \textbf{47.6} & \textbf{37.8} \\
\textbf{Panoptic PartFormer (ResNet101)} & & \textbf{49.2} & \textbf{39.3} \\
\bottomrule
\end{tabular}
\end{adjustbox}
\caption{\textbf{Experiment Results on PPP dataset.} All the methods use single scale inference.}
\label{tab:experiments_res_ppp}
\end{table}

\begin{table*}[!t]
    \footnotesize
	\centering
    \scalebox{0.70}{\subfloat[Effect of each component. DD: Decoupled Decoder. DC: Dynamic Convolution.
    SA: Self Attention. I: Interaction number.
    ]{
        \small
        \label{tab:effect_component}
	    \begin{tabularx}{0.40\textwidth}{c c c c c c c} 
		        				\toprule[0.15em]
    		DD & DC  & SA & I=1 & I=3 & PQ & PartPQ  \\
            \midrule[0.15em]
    		\checkmark & \checkmark & \checkmark &  - & \checkmark & 61.6 & 57.4 \\
    		- & \checkmark & \checkmark & - & \checkmark & 61.2 & 55.9 \\ 
    		\checkmark & - & \checkmark & - & \checkmark & 57.0 & 52.2 \\ 
    		\checkmark & \checkmark & - &- & \checkmark & 57.3 & 53.4 \\ 
    		\checkmark & \checkmark & \checkmark &\checkmark & - & 58.3 & 54.2 \\ 
        	\bottomrule[0.1em]
	    \end{tabularx}
    }} \hfill
    \scalebox{0.70}{
    \subfloat[Ablation on Query Reasoning Design]{
        \label{tab:query_reasoning}
		\begin{tabularx}{0.40\textwidth}{c c c} 
			\toprule[0.15em]
		  Setting & PQ & PartPQ \\
			\midrule[0.15em]
		  Joint Reasoning & 61.6 & 57.4 \\
          Separate Reasoning & 61.1 & 56.8 \\
          Sequential Reasoning & 60.8 & 56.3 \\
			\bottomrule[0.1em]
		\end{tabularx}
    }} \hfill
    \scalebox{0.70}{
    \subfloat[Dense Prediction or Query Prediction on Part. DP: Dense Prediction.
    w: with. ASPP: Atrous Spatial Pyramid Pooling~\cite{deeplabv3}.    
    ]{
        \label{tab:dp_vs_jq}
		\begin{tabularx}{0.45\textwidth}{c c c} 
			\toprule[0.20em]
			Method & PQ & PartPQ \\
			\midrule[0.15em]
			Joint Query  & 61.6 & 57.4 \\ 
            DP-Based & 59.8 & 55.9 \\ 
            DP-Based w ASPP~\cite{deeplabv3} & 59.9 & 56.1 \\ 
			\bottomrule[0.1em]
		\end{tabularx}
    }} \hfill
    \scalebox{0.65}{
    \subfloat[Effect of Aligned Decoder Design.]{
        \label{tab:aligned_decoder}
	    \begin{tabularx}{0.80\textwidth}{c c c c c} 
		        				\toprule[0.15em]
    		 Settings  & PQ & PartPQ & P & NP \\
    		\toprule[0.15em]
    		w Aligned Part Decoder & 61.6 & 57.4 & 43.9 & 62.2\\
    		w/o Aligned Part Decoder & 61.4 & 56.3 & 41.2 & 62.1  \\
    		Aligned Part Decoder on Both Features &  61.4 & 57.2 & 43.7 & 62.0 \\
        	\bottomrule[0.1em]
	    \end{tabularx}
    }} \hfill
    \scalebox{0.70}{
    \subfloat[Upper bound Analysis. GT: Ground Truth]{
        \label{tab:upper_bound}
	    \begin{tabularx}{0.55\textwidth}{c c c c c} 
		  \toprule[0.15em]
    	    Setting & Panoptic-GT  & Part-GT  & PartPQ & P \\
    		\toprule[0.15em]
    	       baseline & - & - & 57.4 & 43.9 \\
    	        & - & \checkmark & 61.6 & 56.1 \\
    	        & \checkmark &  - & 88.4 & 56.4 \\
        	\bottomrule[0.1em]
	    \end{tabularx}
    }} \hfill
	\caption{Ablation studies and analysis on Cityscapes Panoptic Part validation set with ResNet50 as backbone. Best view it in color.}
\end{table*}

\begin{table}[t]
\centering
\begin{adjustbox}{width=0.42\textwidth}
\begin{tabular}{c c c c c}
\toprule[0.15em]
\textbf{Method}  &  Backbone  & $PQ$ & $PQ_{th}$ & $PQ_{st}$  \\ 
\midrule[0.15em]
Panoptic FPN~\cite{kirillov2019panopticfpn} & ResNet101 & 58.1 & 52.0 & 62.5 \\
  UPSNet~\cite{xiong2019upsnet} & ResNet50 & 59.3 & 54.6 & 62.7 \\
  SOGNet~\cite{yang2019sognet} & ResNet50 & 60.0 & { 56.7} & 62.5 \\
  Seamless~\cite{porzi2019seamless} & ResNet50 & 60.2 & 55.6 & 63.6 \\
  Unifying~\cite{li2020unifying} & ResNet50 & 61.4 & 54.7 & 66.3 \\
  Panoptic-DeepLab~\cite{cheng2020panoptic} & ResNet50 & 59.7 & - & - \\
  Panoptic FCN$^*$~\cite{li2020panopticFCN} & ResNet50 & {\ 61.4} & 54.8 & { 66.6} \\
  Panoptic FCN++~\cite{li2021fully} & Swin-large & 64.1 & 55.6 & 70.2 \\
  \hline
Panoptic-PartFormer & ResNet50 & 61.6 & 54.9 & 66.8 \\
Panoptic-PartFormer & Swin-base & 66.6 & 61.7 & 70.3 \\
\bottomrule
\end{tabular}
\end{adjustbox}
\caption{ \small \textbf{Experiment Results on Cityscapes Panoptic validation set.}$^*$ indicates using DCN~\cite{deformablev2}.}
\label{tab:experiments_res_cityscapes}
\end{table}

\subsection{Ablation Study and Model Design}
\label{sec:ablation}
In this section, we present ablation study and several model designs of our Panoptic-PartFormer. 

\noindent
\textbf{Effectiveness of each component.} As shown in Tab.~\ref{tab:effect_component}, we start with the effectiveness of each component of our framework by removing it from the original design. Removing Decoupled Decoder (DD) results in a 1.4 \% drop on PartPQ. Removing Dynamic Convolution (DC) or Self Attention (SA) results in a large drop on PQ which means both are important for the interaction between queries and corresponding query features. Decreasing stage number to 1 also leads to a significant drop. Performing more interaction results in more accurate feature location for each query, which is the same as previous works~\cite{zhang2021knet,peize2020sparse}.

\noindent
\textbf{Whether the part query depends more on thing query?} With our framework, we can easily analyze the relationship among stuff query, thing query, and part query. We present several ways of reasoning and fusing different queries. From intuitive thought, part information is more related to thing query. We design two different query interaction methods, shown in Tab~\ref{tab:query_reasoning}. For separate reasoning, we adopt DD and SA on two query pairs, including stuff-thing query and thing-part query. For sequential reasoning, we perform DD and SA with thing-part query first and stuff-thing query second. However, we find the best model is the joint reasoning, which is the default setting described in method part. We argue that better part segmentation needs the whole scene context rather than thing features only. 

\begin{table*}[!t]
    \footnotesize
	\centering
    \scalebox{0.70}{\subfloat[Effect on Position Encoding.
    ]{
        \small
        \label{tab:pos_enc}
	    \begin{tabularx}{0.45\textwidth}{c c c} 
		 	\toprule[0.2em]
				Method & PQ & PartPQ \\
				\midrule[0.15em]
				baseline &  61.6  & 57.4 \\
				w/o positional encoding & 59.0  & 55.1  \\
				\bottomrule[0.1em]
	    \end{tabularx}
    }} \hfill
    \scalebox{0.70}{
    \subfloat[Effect of adding boundary supervision]{
        \label{tab:boundary_loss}
		\begin{tabularx}{0.40\textwidth}{c c c} 
		\toprule[0.2em]
				Method &  PQ  & PartPQ  \\  
				\midrule[0.15em]
				baseline &  61.6  & 57.4 \\
			    + boundary loss & 61.5  & 57.2  \\
			\bottomrule[0.1em]
		\end{tabularx}
    }} \hfill
    \scalebox{0.70}{
    \subfloat[Effect on adding part annotations.]{
        \label{tab:part_anno}
		\begin{tabularx}{0.45\textwidth}{c c c} 
		\toprule[0.2em]
				Method &  PQ  & PartPQ  \\  
				\midrule[0.15em]
				things and stuff only  &  61.2  & - \\
				+ part annotation (ours) & 61.6  & 57.4  \\
			    \bottomrule[0.1em]
		\end{tabularx}
    }} \hfill
     \caption{\small More analysis using our Panoptic-PartFormer.}
\end{table*}

\noindent
\textbf{Choose joint query modeling or separate modeling on part?} Following PanopticFPN settings~\cite{kirillov2019panopticfpn}, we also adopt semantic-FPN like model for part segmentation. Dense Prediction (DP) is the baseline method shown in Fig.~\ref{fig:teaser_01}(b). We adopt the same merging process for panoptic segmentation and part segmentation. As shown in Tab.~\ref{tab:dp_vs_jq}, our joint query based method achieves the better results and outperforms previous dense prediction based approach and its improved version~\cite{deeplabv3}. This indicates that joint learning benefits part segmentation a lot, which proves the effectiveness of our framework.  

\noindent
\textbf{Aligned decoder is more important for part segmentation.} As shown in Tab.~\ref{tab:aligned_decoder}, using aligned part decoder results in better PartPQ especially for the things with Part (P). Adding both paths with aligned decoder does not bring extra gain. This verifies our motivation: part segmentation needs more detailed information, while thing and stuff predictions do not need it.

\begin{figure*}[t!]
	\centering
	\includegraphics[width=0.75\linewidth]{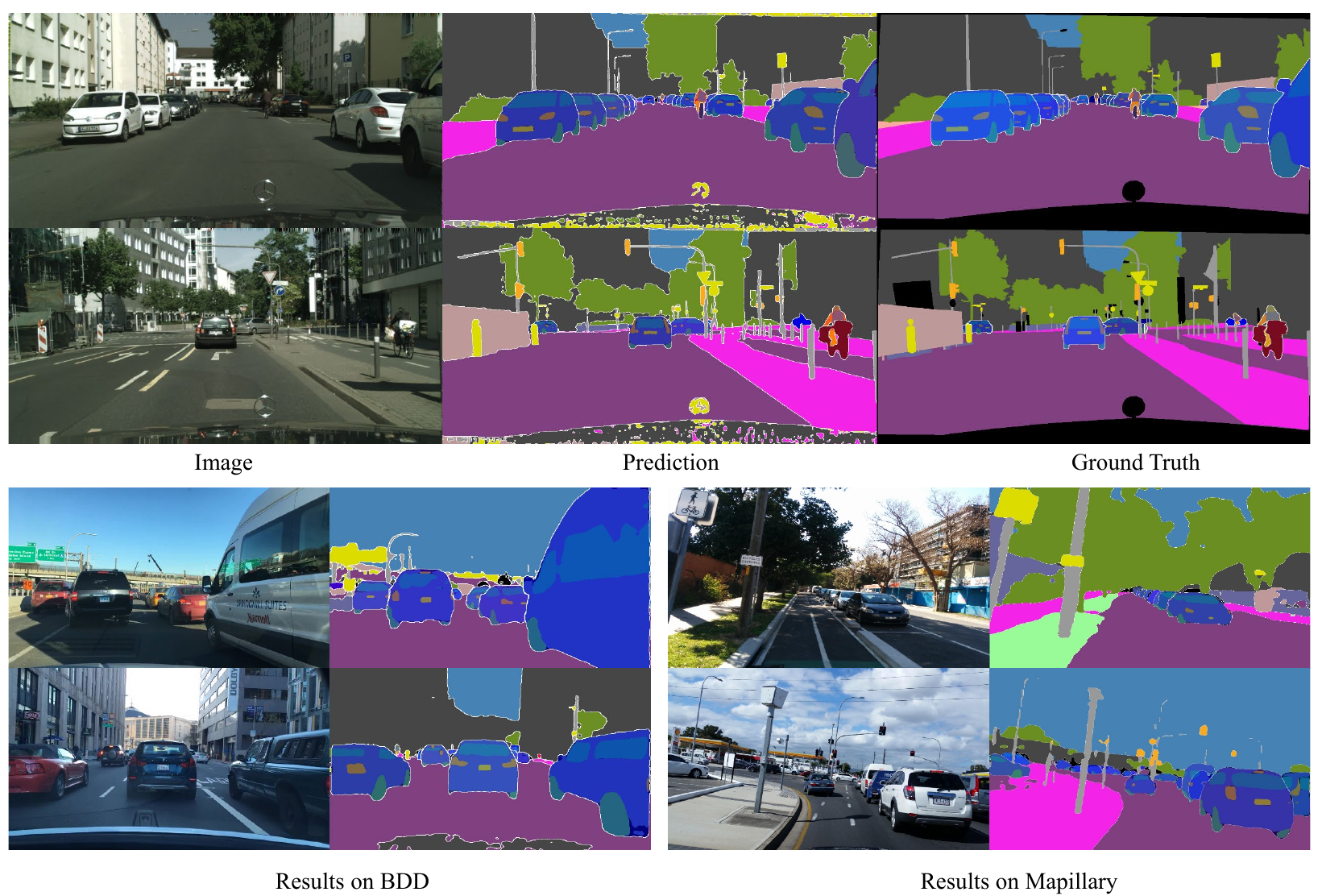}
	\caption{\small Visualization of our Panoptic-PartFormer. Top: results on Cityscapes PPS validation set. Bottom left: prediction on BDD dataset~\cite{yu2020bdd100k}. Bottom right: prediction on Mapillary dataset~\cite{neuhold2017mapillary}. Best view it on screen. }
	\label{fig:exp_vis}
\end{figure*}

\noindent
\textbf{Necessity of Positional Encoding on $X_{p}$ and $X_{u}$.} 
In Tab.~\ref{tab:pos_enc}, removing positional encoding leads to inferior results on both PQ and PartPQ which indicates the importance of position information~\cite{wang2020solov2,zhang2021knet,detr}.

\noindent
\textbf{Will boundary supervision help for part segmentation?} In Tab.~\ref{tab:boundary_loss}, we also add boundary supervision for part segmentation where we use the dice loss~\cite{milletari2016v} and binary cross entropy loss. However, we find no gains on this since our mask is generated from aligned decoder since it already contains detailed information.

\noindent
\textbf{Will joint training help for panoptic segmentation?} As shown in Tab.~\ref{tab:part_anno}, joint learning benefits the panoptic segmentation baseline. However, the benefit is \textit{limited} since both thing and stuff prediction does not need much detailed information. 

\subsection{Analysis and Visualization}

\begin{figure*}[!t]
	\centering
	\includegraphics[width=0.75\linewidth]{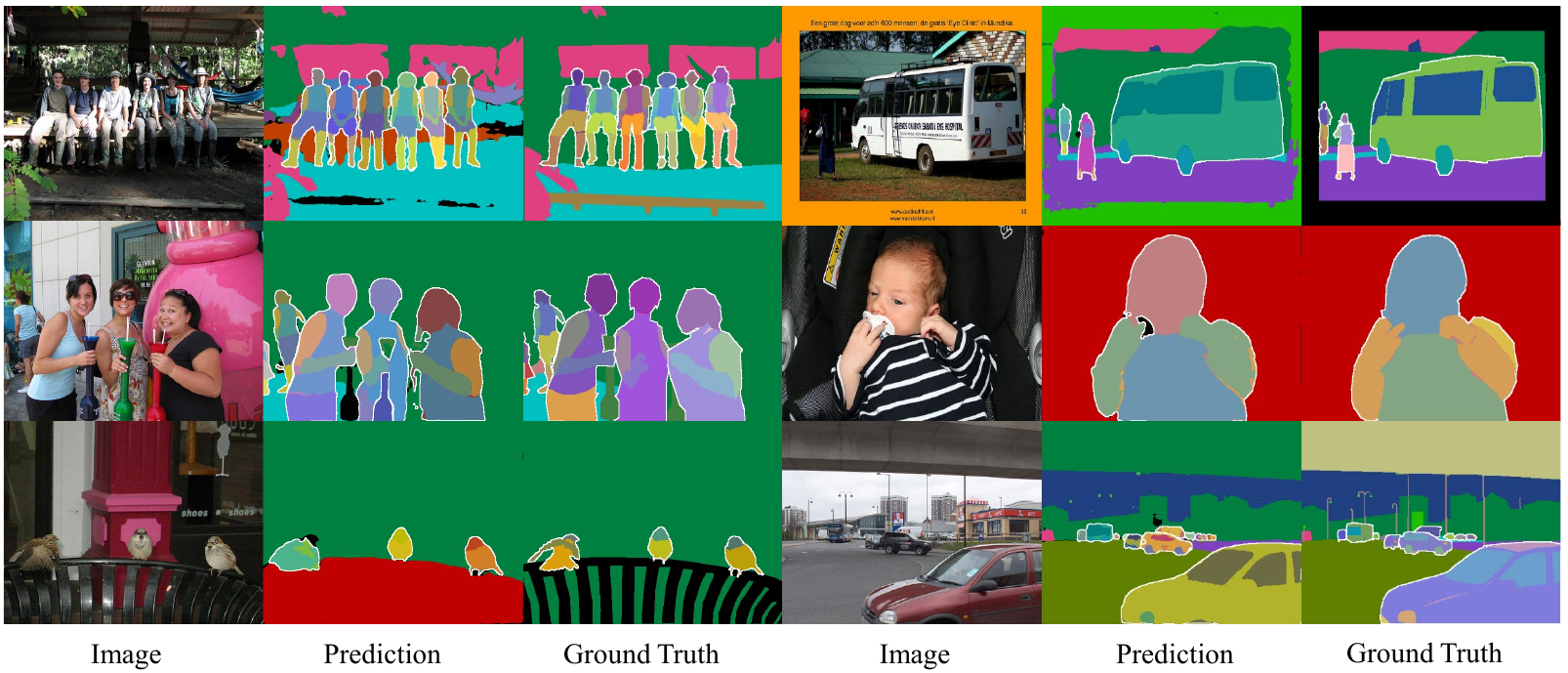}
	\caption{\small More visualization results on Pascal Context Panoptic Part validation set. Best viewed in color and by zooming in. Note that stuff classes have the same color, while thing classes are not.}
	\label{fig:vis_results_pascal_pss}
\end{figure*}
\noindent
\textbf{Visualization and Generalization.} We give several visualization examples using our model on Cityscapes PPS validation set. Moreover, we also visualize several examples on the Mapillary dataset~\cite{neuhold2017mapillary} and BDD dataset~\cite{yu2020bdd100k} to show the generalization ability of our method. As shown in the first row of Fig.~\ref{fig:exp_vis}, our method achieves considerable results. Moreover, on the Mapillary~\cite{neuhold2017mapillary} and BDD datasets~\cite{yu2020bdd100k} which do not have part annotations, our method can still work well as shown in the last row of the Fig.~\ref{fig:exp_vis}. Moreover, we also visualize the results on PPP datasets. The first two rows show the crowded human scene and outdoor scene. Both cases show that our model can obtain the convincing results. The last row shows the small objects cases. The failure cases are due to tiny objects including their parts. 

\noindent
\textbf{Upper bound analysis of Our Model.} In Tab.~\ref{tab:upper_bound}, we give the upper bound analysis to our model by replacing the panoptic segmentation Ground Truth or part segmentation Ground Truth into our prediction. Replacing panoptic segmentation GT leads to a huge gain on PartPQ while replacing part segmentation only results in a limited gain. That indicates PartPQ is more \textit{sensitive to panoptic segmentationt than part segmentation on CPP dataset}. We conclude that a stronger panoptic segmentation model maybe the key for better PPS results. 


\section{Conclusion}
\label{sec:conclusion}
In this work, we present Panoptic-PartFormer, the first unified end-to-end model for Panoptic Part Segmentation. We present a decoupled decoder with three different queries to generate thing, stuff and part masks at the same time. We propose to jointly learn the three queries with corresponding query features. With this framework, we present detailed analysis of the relationship among things, stuff, and part. As a result, our method achieves the new state-of-the-art results on Cityscapes Panoptic PPS dataset and Pascal Contexct PPS dataset. Panoptic-PartFormer would serve as a unified baseline and benefit multiple level concepts scene understanding by easing the idea development. 


\noindent
\textbf{Acknowledgement.} This research is also supported by the National Key Research and Development Program of China under Grant No. 2020YFB2103402. We thank the computation resource provided by SenseTime Research.


\clearpage
%
%
\bibliographystyle{splncs04}
\bibliography{egbib}
\end{document}